%% file: main.tex
\documentclass[runningheads]{llncs}

\usepackage{amsmath}
\usepackage{array}
\usepackage{booktabs}
\usepackage{graphicx}
\usepackage{hyperref}
\usepackage[protrusion=true,expansion=false]{microtype}
\usepackage{url}
\usepackage{tikz}
\usetikzlibrary{arrows.meta,positioning,fit,backgrounds,shapes.geometric,calc}
\usepackage{placeins}
\usepackage{pifont}
\usepackage{xcolor}
\usepackage[normalem]{ulem}
\graphicspath{{figures/}}

\newcommand{\finaltext}[1]{#1}
\newcommand{\removedtext}[1]{}
\newcommand{\finalsep}{}

\makeatletter
\renewcommand\paragraph{\@startsection{paragraph}{4}{\z@}%
                       {-12\p@ \@plus -4\p@ \@minus -4\p@}%
                       {-0.5em \@plus -0.22em \@minus -0.1em}%
                       {\normalfont\normalsize\bfseries}}
\makeatother

\newcolumntype{L}[1]{>{\raggedright\arraybackslash}p{#1}}

\title{Option-Aware Retrieval and Task-Specific VLM Adaptation for Medical VQA}
\titlerunning{Option-Aware Retrieval and Task-Specific Adaptation for Medical VQA}

\author{\removedtext{}\finalsep\finaltext{Tristan Kirscher}\inst{1,2,*} \and
\finaltext{Niklas C. Koser}\inst{3,*} \and
\finaltext{Soren Pirk}\inst{3}}
\authorrunning{\removedtext{}\finalsep\finaltext{T. Kirscher and N. C. Koser et al.}}
\institute{\removedtext{}\finalsep\finaltext{ICube Laboratory, CNRS UMR 7357, University of Strasbourg,
Strasbourg, France} \and
\finaltext{CLCC Institut Strauss, Strasbourg, France} \and
\finaltext{Visual Computing and Artificial Intelligence, Kiel University, Kiel, Germany}\\
\finaltext{\textsuperscript{*}Contributed equally.}}

\begin{document}
\maketitle

\begin{abstract}
We describe our submission to the MedReason 2026 challenge, covering multiple-choice (MCQ) and open-ended (OE) medical visual question answering (VQA) under fully offline, containerized inference. 
Our first finding is that MCQ retrieval must compare answer \emph{semantics} rather than answer labels: labels are independently assigned per question, so copying a retrieved neighbor's label transfers no useful information, whereas scoring each current option's text against correct-answer text from similar training cases raises retrieval-only accuracy from 20.0\% to 57.5\% on a 200-case retrieval-excluded development holdout. 
Our second finding attributes the submitted system's accuracy: holding the task-specific MCQ Low-Rank Adaptation (LoRA) adapter fixed and varying the number \(k\) of in-prompt retrieved examples changes accuracy by at most one case --- 187/200 (93.5\%) at both \(k=0\) and the adapter's training-time \(k=1\), 188/200 (94.0\%) at the packaged runtime's default \(k=3\) --- and the submitted confidence-gated override adds no net accuracy on top of \(k=3\), selecting the VLM in 198/200 cases. 
\removedtext{}
\finaltext{With the final MCQ adapter fixed, retrieval changes accuracy by at most one case, and gating provides no net gain.}
\removedtext{}
\finaltext{On 20 OE cases, token-F1 and RaTEScore~\cite{zhao2024ratescore} decrease as \(k\) grows, but paired sign tests on token-F1 differences are nonsignificant (\(p \ge 0.29\)); a single-annotator comparison found 6/20 wrong-anchor errors for the final configuration and 14/20 for an earlier configuration that jointly differed in routing, adapter, and prompting.}
The system reaches 94.0\% MCQ accuracy on the development holdout and 93.20\% on the organizer's official pre-evaluation, versus 29.43\% for the off-the-shelf reference baseline, while both of the organizer's open-ended scores are lower than that baseline's (ground-truth agreement 1.245 versus 1.588, visual accuracy 1.995 versus 2.696, each out of 4).
\removedtext{}
\finaltext{Code and configuration are available at \url{https://github.com/Kirscher/MedReason2026}.}

\end{abstract}

\section{Introduction}
\label{sec:intro}
This paper describes our submission to the MedReason 2026 challenge~\cite{medreason2026challenge}, a medical visual question answering (VQA) benchmark spanning multiple-choice (MCQ) and open-ended (OE) questions under a fully offline, containerized inference constraint: the system receives no network access and must run end-to-end inside a packaged Docker image. 
MCQ cases require selecting one official option label, OE cases require a concise answer and a reasoning trace. Recent vision-language models (VLMs) provide a strong multimodal foundation, but answers must still be image-grounded, semantically precise, and format-constrained.

Retrieval augmentation is attractive here because similar training cases can supply task-specific terminology and answer priors.
Nearest-neighbor label transfer, however, is structurally wrong for MCQ: option labels are local to each question, so option \texttt{A} in one case has no semantic relationship to option \texttt{A} in another, and copying the nearest example's label can be confidently wrong even when that case is topically relevant. 
The quantity to compare is the candidate answer content, not the arbitrary letter assigned to it, and we show that treating retrieval as an option-aware semantic prior corrects most of this failure mode.

We additionally test whether retrieval confidence can identify cases in which a retrieval prediction should replace the adapted VLM prediction. In practice, this gate is nearly inactive, and our fixed-adapter ablation shows that it provides no net accuracy gain (Sec.~\ref{sec:results}). We therefore report it as a negative result rather than as a principal source of performance.

Our contributions are: (1) we quantify a structural failure of nearest-label retrieval for MCQ-VQA and show that scoring the semantic content of the current options is substantially more effective (20.0\% to 57.5\%); \removedtext{}\finaltext{(2) we show that retrieval and gating provide at most a one-case MCQ gain with the adapter fixed; and (3) we report preliminary OE results from 20 cases.}

\section{Related Work}
\label{sec:related}

\paragraph{Medical VQA and Vision-Language Adaptation.}
Medical VQA datasets such as VQA-RAD~\cite{lau2018vqarad}, SLAKE~\cite{liu2021slake}, and PMC-VQA~\cite{zhang2023pmcvqa} differ from generic VQA in vocabulary, modality heterogeneity, and required answer precision. 
\finaltext{Med-CMR evaluates fine-grained visual evidence and clinical reasoning in complex medical questions~\cite{gong2026medcmr}.}
General-purpose VLMs provide strong multimodal priors~\cite{bai2025qwen25vl}, and biomedical variants such as LLaVA-Med~\cite{li2023llavamed} further specialize them, but adaptation still matters for a specific challenge dataset and output contract.
Parameter-efficient fine-tuning (PEFT) makes such adaptation practical: we use LoRA~\cite{hu2022lora} over a 4-bit quantized backbone following the QLoRA recipe~\cite{dettmers2023qlora}, with separate adapters for MCQ and OE generation.

\paragraph{Retrieval and Confidence-Based Routing.}
Retrieval-augmented generation uses external evidence to improve knowledge-intensive prediction~\cite{lewis2020rag}, including medical QA benchmarks~\cite{xiong2024medrag}. Our method differs from nearest-neighbor label transfer by scoring retrieval evidence against each current candidate option's text, i.e.\ using retrieval as an option-level semantic prior, which avoids 
the question-local label problem described above. 
Selective prediction and confidence-based routing defer between predictors on a scalar signal, justified by an assumed
correlation between confidence and correctness that calibration research finds is often weaker than supposed~\cite{guo2017calibration}. 
We use such a rule to arbitrate between the retrieval prior and the VLM, without calibrating the VLM's probabilities.

\FloatBarrier
\section{Method}
\label{sec:method}
\begin{figure}[t]
\centering
\includegraphics[width=\textwidth]{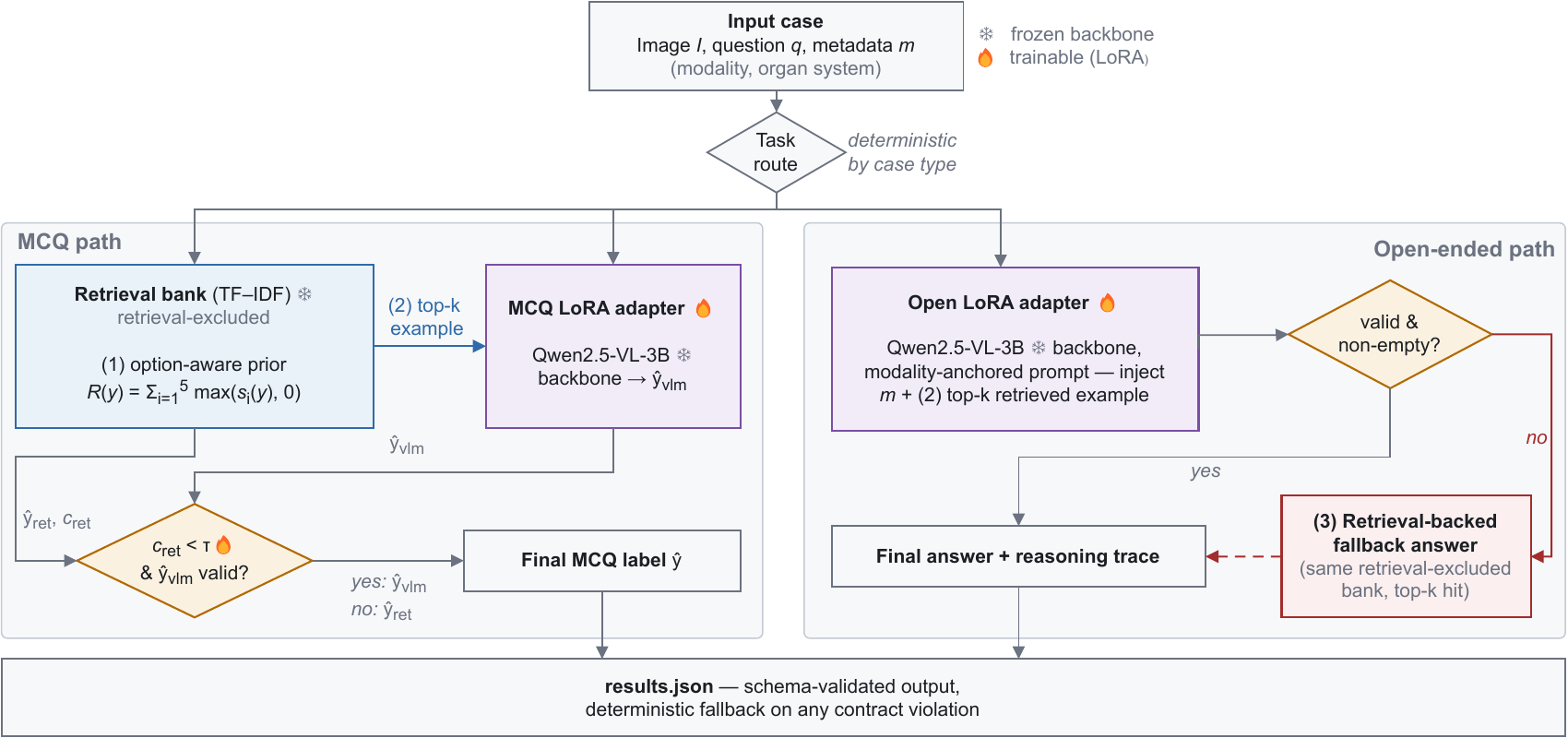}
\caption{Inference pipeline. Task type is dispatched deterministically, and one retrieval-excluded TF--IDF bank is used three ways: (1) an option-aware decision prior $\hat y_{ret}, c_{ret}$ for the MCQ confidence gate; (2) $k$ in-prompt retrieved examples for both adapted VLMs ($k=1$ during adapter training, $k=3$ in the submitted runtime); (3) a retrieval-backed fallback answer (dashed), only when OE generation violates the output contract.}
\label{fig:overview} 
\end{figure}

Figure~\ref{fig:overview} summarizes the inference pipeline.
Prompts, the term frequency and inverse document frequency (TF--IDF) retrieval, and per-fold threshold values will be released with the code.


\paragraph{Problem Formulation.}
Each case consists of an image \(I\), a question \(q\), and metadata \(m\) (modality, organ system, task, subtype). 
MCQ cases carry a finite option set \(Y=\{(l_1,t_1),\ldots,(l_K,t_K)\}\) of labels \(l_j\) and texts \(t_j\) for \(j=1,\ldots, K\), from which exactly one label must be returned. OE cases require a concise answer plus a non-empty reasoning trace.

\paragraph{Offline Backbone and Task-Specific Adapters.}
The multimodal backbone is a frozen Qwen2.5-VL-3B-Instruct, with two LoRA adapters, one per task format, both trained with exactly one retrieved example in the training prompt (\(k=1\)). 
The packaged runtime leaves the retrieved-example count unset and therefore falls back to the code default \(k=3\), 
and this mismatch between training and runtime is quantified in Table~\ref{tab:ablation}, rows 6--9.
Routing is deterministic dispatch on the declared case type, rather than a learned router, to avoid applying the MCQ-specialized adapter to free-text generation, where informal checks showed degraded output.
The final MCQ adapter was selected through a bundled training-alignment sweep that aligns the fine-tuning and runtime MCQ prompt formats, except for the retrieval-count mismatch. 
Training additionally uses deterministic answer-option permutations and an answer-first JSON target to improve parsing reliability. 

\paragraph{Training and Inference Configuration.}
Both adapters share a single PEFT LoRA setting (rank \(r=16\), \(\alpha=32\), dropout \(0.05\), no bias) applied to all attention and MLP projections.  
Base weights are loaded in 4-bit NormalFloat (NF4) with double quantization and fp16 compute during training; LoRA weights are trained in full precision, and inference uses the unquantized backbone. Both adapters use AdamW at \(1\times10^{-4}\) with an effective batch size of 4, for 250 steps over 4{,}096 option-permuted MCQ examples and 200 steps over 1{,}024 OE examples, respectively, and both training sets exclude every development-holdout case identifier. 
Images pass unmodified through the default Qwen2.5-VL dynamic-resolution processor, and inference is greedy with a 384-token budget, so runs at a fixed configuration are byte-identical.

\paragraph{Retrieval Bank and Option-Aware Prior.}
The retrieval bank is built from released training cases; each entry holds question text, answer text, visual description, options, and metadata. 
In every reported experiment, case identifiers from the evaluated holdout are removed before the bank is built, since retrieval can otherwise memorize exact cases rather than generalize across similar ones. 
Retrieval uses TF--IDF, which is offline, deterministic, and inexpensive 
to package.
The prior is purely textual, and the query never includes the current image. For each candidate option \(y=(l,t)\) we query \(q\), option text \(t\), and \(m\) against documents built from training questions, visual descriptions, metadata, and correct-answer text, where those descriptions are captions rather than image features. 
If \(s_i(y)\) is the similarity of the \(i\)-th of the five highest-scoring documents for option \(y\), the option-aware score is
\begin{equation}
  R(y)=\sum_{i=1}^{5}\max(s_i(y),0),
  \qquad
  \hat{y}_{ret}=\arg\max_{y\in Y} R(y).
\end{equation}

\paragraph{Confidence-Gated VLM Override.}
The adapted VLM predicts an option label \(\hat{y}_{\mathrm{vlm}}\), while retrieval generates an alternative prediction \(\hat{y}_{\mathrm{ret}}\) together with a confidence score \(c_{\mathrm{ret}}\). 
The two predictions are combined through a gating rule that retains \(\hat{y}_{\mathrm{vlm}}\) when it is valid and \(c_{\mathrm{ret}}\) stays below a threshold \(\tau\), and selects \(\hat{y}_{\mathrm{ret}}\) otherwise. 
Here, \(c_{\mathrm{ret}}\) denotes a heuristic retrieval-confidence score, defined as the highest option-aware score divided by the sum of the scores across all current answer options. It should therefore not be interpreted as a calibrated probability.
The operating threshold \(\tau=0.257\) is derived from the fold protocol described in Sec.~\ref{sec:experiments}.
Applied to the 200 MCQ cases of H220, the gating rule retains the VLM prediction in almost all instances. The retrieval-confidence score averages \(0.223\), ranging from \(0.201\) to \(0.260\), and falls below the threshold of \(\tau=0.257\) in 198 out of 200 cases (99.0\%). 
The retrieval prediction is therefore returned in only two cases, and pooled accuracy is identical with and without the gate (Sec.~\ref{sec:results}), so the conclusions do not depend on the threshold choice.


\paragraph{Open-Ended Routing, Anchoring, and Validation.}
OE cases are processed using the OE adapter. 
When available, the imaging modality and anatomical region are included in both training and inference prompts. 
The model is instructed to ground its response in this information and disregard retrieved examples from other modalities or anatomical regions. 
The released visual descriptions are used as reasoning-trace targets rather than generic placeholders. 
The separate OE path was introduced following a single-annotator review of 20 OE outputs generated with the MCQ-specialized adapter. 
All 20 were judged qualitatively imperfect, and 14 referred to an unrelated modality or anatomical region. 
These 14 outputs were classified as \emph{wrong-anchor} errors. 
A comparable annotation of the final OE outputs is reported in Sec.~\ref{sec:results}. 
Empty, placeholder, or malformed answer fields trigger a compact retrieval-backed fallback. This ensures a valid output format but is not claimed to provide a clinically sufficient answer.

\section{Experiments}
\label{sec:experiments}
\paragraph{Data and Splits.}
All experiments use labeled holdouts sampled from the organizer's released MedReason 2026 training package (17{,}722 train / 2{,}532 validation cases, stratified by question type, task, and organ system; every MCQ case has \(K=5\) options). 
We evaluate on a 220-case holdout (200 MCQ, 20 OE; the primary analysis split, denoted H220) and a 60-case holdout (50 MCQ, 10 OE; H60), independently sampled and nearly disjoint (1 shared case). Holdout identifiers are excluded from the retrieval bank in every reported experiment. 
\removedtext{}
\finaltext{These are development holdouts, not independent test sets. H220 was reused for adapter, prompt, threshold, and post-hoc retrieval-field selection; the OOF analysis addresses threshold selection only. Case-ID exclusion does not prevent image overlap, which we audit in Sec.~\ref{sec:discussion}.}

\paragraph{Metrics.}
For MCQ cases, exact label accuracy and a 5-fold out-of-fold (OOF) estimate are reported. Each case is assigned to one of five folds by hashing its identifier, \(\tau\) is selected on the other four by grid search over \([0.15,0.35]\) in steps of \(0.001\), and the held-out predictions are pooled, so no case is used to select the threshold under which it is evaluated. 
The threshold in the submitted container, \(\tau=0.257\), is the median of the five fold-specific values, whereas optimization on all 200 cases yields \(\tau=0.241\); both depend on labels from the full holdout and are therefore optimistic. 
The OOF estimate accounts only for threshold selection, not for the prior selection of the adapter, prompt, or architecture on the same holdout. 
For OE cases, lexical token-F1 and RaTEScore~\cite{zhao2024ratescore}, an entity-aware and negation-sensitive metric for radiology text similarity, are reported; neither measures clinical correctness or visual grounding.

\section{Results}
\label{sec:results}

\input{results}

Table~\ref{tab:ablation} reports MCQ accuracy on the 200 MCQ cases of H220 and OE scores on its 20 OE cases. 
Nearest-label transfer reaches 40/200 --- exactly the 20\% chance rate for \(K=5\). 
Comparing answer semantics 
corrects this failure, and our particular formulation is not required for it.
Matching option text directly against the correct-answer text of retrieved cases, with no question or metadata on either side, already reaches 115/200 (row 2), the same total as the submitted prior (row 3), although the two agree on only 121/200 cases.
Which fields enter the comparison has a larger effect than we anticipated.
Over a 24-point grid of index, query, and aggregation variants, the best configuration, option text plus metadata against answer text plus metadata summed over 5 hits, reaches 149/200 (74.5\%).
The submitted prior is thus not the strongest available, although that grid was scanned on H220 itself and the gate keeps a retrieval answer in only 2/200 cases, so the submitted system is nearly insensitive to which retrieval-field configuration is used.
Every semantic variant 
exceeds the unadapted backbone (row 4), and nearest-label transfer does not. 
\removedtext{}
\finaltext{Row 5 was selected from 20 system candidates; nine OE adapter/prompt variants were also evaluated on H220.}
Rows 6--9 use a further-refined MCQ adapter from a \emph{bundled} training-alignment sweep, which introduced runtime-style prompting, an in-training retrieved example, option-permutation augmentation, answer-first targets, and cosine scheduling, introduced together, 
so we do not attribute the gain from 170/200 to 188/200 to any one of them.

\paragraph{Contributions of Retrieval and of the Confidence Gate.}
The bottom block isolates retrieval content from the confidence gate. 
Both \(k=0\) (row 6) and \(k=1\) (row 7, the adapter's training-time count) reach 187/200, so restoring exactly the training-time retrieved-example count does not change the pooled accuracy.
Only \(k=3\) (row 8, the runtime default) reaches 188/200, one case better, and it does so by supplying \emph{more} context than training used rather than by matching it.
Adding the confidence gate on top of \(k=3\) (row 9, the submitted system) yields \emph{no net accuracy gain} since the gate selects the VLM in 198/200 cases and the two retrieval overrides leave the pooled score unchanged at 188/200.
\finaltext{This ablation isolates inference-time retrieval and gating, not adapter development or visual grounding.}
On H60 the submitted system reaches 47/50 (94.0\%), and the OOF estimate on H220 gives 185/200 (92.5\%).
\removedtext{}
\finaltext{The organizer's pre-evaluation reports 93.20\% MCQ accuracy, open-ended ground-truth agreement of 1.245/4, and open-ended visual accuracy of 1.995/4~\cite{medreason2026leaderboard}.}
The corresponding values for the organizer's off-the-shelf Qwen2.5-VL-7B reference baseline are 29.43\%, 1.588, and 2.696, so our system reaches substantially higher MCQ accuracy while both open-ended scores are lower, a marked discrepancy between closed-ended and open-ended performance. This pattern is not specific to our system, since 15 of 16 ranked teams exceeded the baseline's MCQ accuracy while only one exceeded its visual accuracy.

\paragraph{Open-Ended Results.}
The OE numbers involve two distinct comparisons.
Relative to the initial task-routed baseline (row 5), both proxies improve in the final system, with token-F1 rising from 0.123 $\to$ 0.151 and RaTEScore from 0.338 $\to$ 0.385.
Within the fixed-adapter ablation (rows 6--9), however, both \emph{decrease} monotonically as \(k\) grows, with token-F1 at 0.212, 0.177, and 0.151 and RaTEScore at 0.401, 0.391, and 0.385 for \(k=0,1,3\).
\removedtext{}
\finaltext{The paired sign tests are not significant (\(p=0.29\) and \(p=0.39\)); we therefore treat the trend as descriptive.}

\begin{figure}[t]
    \centering
    \includegraphics[width=1.0\linewidth]{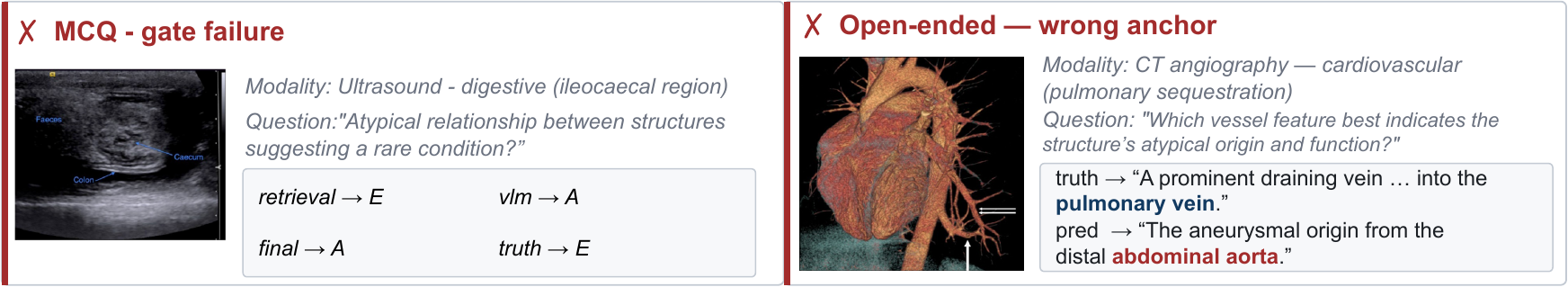}
\caption{Representative MCQ gate failure (left) and OE wrong-anchor failure
(right) from the final packaged Docker run on H220; discussed in the text.}
\label{fig:qual}
\end{figure}

\paragraph{Qualitative Cases.}
Figure~\ref{fig:qual} shows two failure cases from the final packaged Docker
run.
In the first, a gate failure, low retrieval confidence 
assigns the case to the VLM and discards a retrieval prior that was correct. 
In the second, a wrong-anchor OE failure,  the modality is right, but the described anatomy is not. 
\removedtext{}
\finaltext{A single annotator found 6/20 wrong-anchor errors for the final system and 14/20 for the earlier configuration. Because routing, adapter, and prompting differ, this comparison isolates no individual component.}


\FloatBarrier

\section{Discussion and Conclusion}
\label{sec:discussion}
MCQ answer labels are local identifiers, so retrieval for medical MCQ-VQA must compare answer semantics rather than labels, and doing so nearly triples retrieval-only accuracy, independently of LoRA adaptation. 
Nearest-label transfer is a diagnostic that exposes the structural problem rather than the best available retrieval method, and neither is our submitted prior. 
Restricting the comparison to answer text and metadata, with no question or visual-description text, raises retrieval-only accuracy to 149/200 post-hoc. 
\removedtext{}
\finaltext{With the MCQ adapter fixed, \(k=0\) and \(k=1\) reach 187/200, \(k=3\) reaches 188/200, and gating adds no gain. This comparison does not establish visual grounding.}

The OE proxies move in the opposite direction: both fall as \(k\) grows, and the organizer OE scores remain below the off-the-shelf Qwen2.5-VL-7B reference despite substantially higher MCQ accuracy.
This suggests a possible retrieval--grounding trade-off: additional context may provide useful terminology but also anchor the VLM to irrelevant anatomy or modality.
With only 20 cases and nonsignificant paired tests, this remains a hypothesis rather than an established effect.
Moreover, the wrong-anchor reduction from 14/20 to 6/20 was measured against a different baseline and cannot be causally linked to the \(k\) trend.

Several limitations qualify these findings. 
The development holdouts were reused for the selection of adapters, prompts, the threshold, and the retrieval fields behind the 74.5\% figure, and the out-of-fold score estimates \emph{threshold}-selection risk only.
An image-hash audit found 13/200 MCQ holdout cases that share a byte-identical image with a bank case under a
different identifier.
Both retrieval and the adapter reach higher accuracy on this subset than on the remainder, but excluding it changes every reported MCQ accuracy by \(\le1.4\) points without altering any qualitative conclusion.
The TF--IDF bank can miss visually similar cases with weak lexical overlap, the gate threshold is a heuristic rather than a calibrated probability, and neither proxy metric evaluates visual grounding or clinical correctness, so the OE evidence remains preliminary.
\finaltext{Future work should evaluate dense or multimodal retrieval, modality-aware filtering, and image-ablation controls.}
Finally, these conclusions come from one challenge dataset under one offline inference contract, and we do not claim
they generalize beyond MedReason without further evidence.

\begin{credits}
\subsubsection{\ackname}
Data used in this publication were obtained through the MedReason 2026
Challenge project on Synapse (syn74403682)~\cite{medreason2026challenge} and
are governed by the Creative Commons Attribution-NonCommercial 4.0
International license
(\url{https://creativecommons.org/licenses/by-nc/4.0/}). The
participant-facing benchmark builds on
Med-CMR~\cite{gong2026medcmr}, whose images, captions, and metadata were
collected from clinical case reports and research articles published in
biomedical journals, including the \emph{Journal of Medical Case Reports} and
the \emph{New England Journal of Medicine}. The authors thank the MedReason
2026 organizers and contributors.
This work was supported by the IDIR-Project (Digital Implant Research), a
cooperation financed by Kiel University, University Hospital Schleswig-Holstein
and Helmholtz Zentrum Hereon.
This work of the Interdisciplinary Thematic Institute HealthTech, as part of
the ITI 2021--2028 program of the University of Strasbourg, CNRS, and Inserm,
was partially supported by IdEx Unistra (ANR-10-IDEX-0002) and SFRI (STRAT'US
project, ANR-20-SFRI-0012) under the framework of the French Investments for
the Future Program. The authors acknowledge the High Performance Computing
Center of the University of Strasbourg for scientific support and access to
computing resources. Part of the computing resources was funded by the
Equipex Equip@Meso project (Programme Investissements d'Avenir) and the CPER
Alsacalcul/Big Data.

\subsubsection{\discintname}
The authors have no competing interests to declare that are relevant to the
content of this article.
\end{credits}

\newpage

\bibliographystyle{splncs04}
\bibliography{biblio}

\end{document}

%% file: results.tex

\begin{table}[h]
\centering
\caption{Performance on H220 (200 MCQ, 20 OE): MCQ accuracy and OE token-F1/RaTEScore~\cite{zhao2024ratescore}. Rows 2--3 have no OE result because option scoring requires candidate answers; row 1 uses the nearest retrieved answer for OE. \removedtext{}\finaltext{Rows 6--9 keep the task-specific adapters fixed and vary only $k$, isolating inference-time context and gating.} Predictions are deterministic across reruns. The $\pm$ values are bootstrap standard deviations (10{,}000 resamples), not confidence intervals. \textbf{Bold} indicates the best point estimate per column.}
\label{tab:ablation}
\small
\resizebox{\textwidth}{!}{%
\begin{tabular}{lrrr}
\toprule
\textbf{Variant} & \textbf{MCQ accuracy} & \textbf{OE token-F1} & \textbf{OE RaTEScore} \\
\midrule
\multicolumn{4}{l}{\emph{Retrieval and backbone baselines (unadapted)}} \\
\removedtext{}\finaltext{1. Nearest-label retrieval (diagnostic)} & 40/200 (20.0\%) & 0.053 & 0.323 \\
2. Answer-text $\leftrightarrow$ option-text matching & 115/200 (57.5\%) & \multicolumn{2}{c}{n/a} \\
3. Option-aware retrieval (submitted textual prior) & 115/200 (57.5\%) & \multicolumn{2}{c}{n/a} \\
4. Unadapted VLM only (no retrieval, no LoRA) & 50/200 (25.0\%) & 0.041 & 0.345 \\
\midrule
\multicolumn{4}{l}{\emph{Task-specific adaptation}} \\
5. Initial task-routed system (separate MCQ/OE LoRA) & 170/200 (85.0\%) & 0.123 & 0.338 \\
\midrule
\multicolumn{4}{l}{\emph{\removedtext{}\finaltext{Fixed task-specific adapters: inference-time context and gate}}} \\
6. $k=0$: no retrieved example & 187/200 (93.5$\pm$1.7\%) & \textbf{0.212$\pm$0.053} & \textbf{0.401$\pm$0.038} \\
7. $k=1$: training-time count & 187/200 (93.5$\pm$1.7\%) & 0.177$\pm$0.053 & 0.391$\pm$0.041 \\
8. $k=3$: runtime default & \textbf{188/200 (94.0$\pm$1.7\%)} & 0.151$\pm$0.042 & 0.385$\pm$0.046 \\
9. \quad + confidence gate ($k=3$, submitted system) & \textbf{188/200 (94.0$\pm$1.7\%)} & 0.151$\pm$0.042 & 0.385$\pm$0.046 \\
\bottomrule
\end{tabular}%
}
\end{table}